\documentclass[conference]{IEEEtran}
\IEEEoverridecommandlockouts
\usepackage{cite}
\usepackage{amsmath,amssymb,amsfonts}
\usepackage{algorithmic}
\usepackage{graphicx}
\usepackage{textcomp}
\usepackage{xcolor}
\usepackage{multirow}

\usepackage{booktabs}
\usepackage{adjustbox}
\usepackage{svg}
\usepackage{tcolorbox}

\def\BibTeX{{\rm B\kern-.05em{\sc i\kern-.025em b}\kern-.08em
    T\kern-.1667em\lower.7ex\hbox{E}\kern-.125emX}}
\begin{document}

\title{MultiModal-Learning for Predicting Molecular Properties: A Framework Based on Image and Graph Structures\\
}

\author{\IEEEauthorblockN{Zhuoyuan Wang\textsuperscript{1}, 
		Jiacong Mi\textsuperscript{1}, 
		Shan Lu\textsuperscript{2}, 
		Jieyue He\textsuperscript{1,*}}
	\IEEEauthorblockA{
        \textsuperscript{1}\textit{School of Computer Science and Engineering, Key Lab of Computer Network and Information Integration, MOE,}\\ \textit{Southeast University, Nanjing, 210018, Jiangsu, China}\\
		\textsuperscript{2}\textit{Nanjing Fenghuo Tiandi Communication Technology Co., Ltd, Nanjing, 211161, Jiangsu, China}\\
		wangzhuoy@seu.edu.cn, mijiacong@seu.edu.cn, bfcat.cn@gmail.com, jieyuehe@seu.edu.cn}}

\maketitle

\begin{abstract}
The quest for accurate prediction of drug molecule properties poses a fundamental challenge in the realm of Artificial Intelligence Drug Discovery (AIDD). An effective representation of drug molecules emerges as a pivotal component in this pursuit. Contemporary leading-edge research predominantly resorts to self-supervised learning (SSL) techniques to extract meaningful structural representations from large-scale, unlabeled molecular data, subsequently fine-tuning these representations for an array of downstream tasks. However, an inherent shortcoming of these studies lies in their singular reliance on one modality of molecular information, such as molecule image or SMILES representations, thus neglecting the potential complementarity of various molecular modalities. In response to this limitation, we propose MolIG, a novel MultiModaL molecular pre-training framework for predicting molecular properties based on Image and Graph structures. MolIG model innovatively leverages the coherence and correlation between molecule graph and molecule image to execute self-supervised tasks, effectively amalgamating the strengths of both molecular representation forms. This holistic approach allows for the capture of pivotal molecular structural characteristics and high-level semantic information. Upon completion of pre-training,    Graph Neural Network (GNN) Encoder is used for the prediction of downstream tasks. In comparison to advanced baseline models, MolIG exhibits enhanced performance in downstream tasks pertaining to molecular property prediction within benchmark groups such as MoleculeNet Benchmark Group and ADMET Benchmark Group.
\end{abstract}


\section{Introduction}
In the field of computational chemistry and drug discovery, molecular representation learning is a crucial task aimed at developing effective methods to represent molecular structures and predict their properties \cite{goh2017deep,wu2018moleculenet,chen2018rise}. Accurate prediction of chemical molecule properties can greatly facilitate the drug development process, reduce research costs, increase the success rate of new drugs, provide more information and guidance for drug design, and have significant practical value \cite{xiong2019pushing}.
Early methods of molecular representation learning primarily included the topological-based approaches and  physicochemical-based approaches. Topological-based methods describe molecules by analyzing the chemical bond connections between atoms in molecular structures, with the Extended Connectivity Fingerprints (ECFP) \cite{rogers2010extended} being one of the most classic methods. ECFP converts the neighbor information around atoms into fixed-length binary strings as feature representations. On the other hand, physicochemical-based methods describe molecular structures by calculating their physicochemical properties, such as charge states, polarity, electron affinity, ionization energy and etc. These methods require prior calculations of physicochemical properties, which are then used as feature representations. Examples include Molecular Quantum Mechanics (MQM), Molecular Mechanics (MM), Density Functional Theory (DFT) \cite{orio2009density}, etc. However, early methods lacked a deep understanding of molecular structures and the ability to capture complex molecular features, resulting in certain limitations in molecular property prediction tasks.

In recent years, the advent of Graph Neural Networks (GNNs) has brought about remarkable advancements in an array of graph-related tasks \cite{wu2020comprehensive}, subsequently inspiring their application to the learning of molecular structures. Central to the concept of a molecular structure-based GNN model is the perception of the topological structure of atoms and bonds within molecules as a graph, where atoms and chemical bonds correspond to nodes and edges respectively. Initial feature sets are formulated based on their inherent physicochemical properties such as atom type, bond type, and aggregation operations are executed through the iterative exchange of information amongst neighboring nodes \cite{guo2022graph}. In contrast to traditional descriptor-based methods, GNNs can encapsulate a more extensive set of molecular features, including but not limited to local interactions and cyclic structures, thereby enhancing the precision of predictions. To date, numerous GNN-based molecular representation learning methods have been proposed, such as the Message Passing Neural Networks (MPNN) \cite{gilmer2017neural}, and Attentive FP \cite{xiong2019pushing}, among others.

Simultaneously, graph contrastive learning \cite{you2020graph} has been applied to the field of molecular representation learning with the development of GNNs, compensating for the scarcity of labeled molecular data and significantly promoting the development of this field. Existing molecular representation learning methods based on graph contrastive learning usually adopt a graph enhancement strategy for molecules. However, data augmentation strategies in the molecular field are not straightforward due to the specific chemical rules and constraints of molecular structures, which may necessitate additional domain knowledge and experience to design suitable molecular augmentation strategies. In addition to viewing molecules as topological views of nodes and edges, molecules can also be presented in the form of images \cite{shi2019molecular, zhong2021molecular,goh2017chemception}. 
The input of the Graph modality provides the model with detailed information about the molecular structure, such as the types of chemical bonds and atom types. This explicit information transfer enables the model to intuitively understand the microscopic aspects of the molecule. However, it is essential to note that in graph neural networks, each layer attempts to update the feature representation of nodes by aggregating information from their neighbors. This aggregation often leads to high similarity between nodes, resulting in over-smoothing, which limits its expressive power when dealing with complex structures. In this sense, the Graph modality can be considered a form of local modality, emphasizing the expression of microscopic details. In contrast, the input of the Image modality does not provide direct guidance on details like atomic chemical bonds. Although the model is unaware of the fine-grained details of the molecular structure, the global nature of convolutional operations allows each pixel to perceive the overall information of the image. This global perspective helps the model better comprehend the overall context without concerning itself with microscopic details, thereby avoiding the issue of over-smoothing. Therefore, the Image modality can be viewed as a form of global modality, focusing on capturing the overall structure.

Although previous models based on graphs or SMILES have achieved good performance, the information provided by a single modality is limited. Inspired by the significant success of utilizing a large amount of unlabeled data in Computer Vision (CV) and Natural Language Processing (NLP)
, we introduce the molecule image modality into contrastive learning as an augmented modality and propose a novel molecular pretraining framework called MolIG. Considering that molecule graph and molecule image can provide different levels of chemical and geometric information, we capture the consistency between them to capture high-order semantic features of molecules. Compared to previous molecular contrastive learning models \cite{wang2022molecular}, our approach examines molecular features from different perspectives and fully utilizes information from both modalities. Our contributions can be summarized as follows:

\begin{itemize}

\item We propose a pioneering multimodal pre-training model dedicated to molecular property prediction, and train the model using both molecular graphs and molecular images as two modalities.

\item Ensuring the preservation of molecular semantic features, we employ three different image enhancement strategies. By doing so, we maximize the congruence between molecule graph and image modalities, thereby realizing a more robust and universally applicable representation learning.

\item Our model is subjected to evaluation on both the MoleculeNet Benchmark Group and ADMET Benchmark Group. The outcomes of the experiment indicate that our model is proficient not only in extracting the structural attributes of molecules but also in capturing the more elusive higher-order semantic information. Furthermore, it surpasses the state-of-the-art methods that rely on molecule graph and molecule image in terms of results.

\end{itemize}

\section{Relate Work}

\subsection{Graph-Based Molecule Representation}
Graph-based molecular representation learning currently stands as the predominant approach. 
GROVER\cite{rong2020self} ingeniously integrate GNNs and Transformers, generating node embeddings through the prediction of contextual attributes and motif information. Simultaneously, contrastive learning has been extensively employed in the field of molecular representation learning. Conversely, MolCLR \cite{wang2022molecular} adopted a distinct approach by applying random augmentation operations, including Node Masking, Edge Deletion, and Subgraph Removal, to the molecule graph. KANO \cite{fang2023knowledge} utilizes knowledge graph to guide molecular graph enhancement. However, these operations inevitably modify the molecular structure, thus conflicting with established chemical principles.

\subsection{Image-Based Molecule Representation}
ADMET-CNN \cite{shi2019molecular} successfully established a Convolutional Neural Network (CNN) model predicated on two-dimensional molecule image, yielding excellent outcomes in forecasting ADMET attributes. Representation learning methodologies based on molecule image mandate the conversion of data samples into Euclidean space. Yet, due to the absence of attributes related to atoms and bonds, direct predictions of properties using molecule image have been less than satisfactory.

\subsection{MultiModal-Based Molecule Representation}
GeomGCL \cite{li2022geomgcl} has designed a Geometry Message Passing Network that capitalizes on both 2D and 3D dual views, adaptively harnessing the vast information available from both 2D and 3D graphs. 
MoleculeSTM \cite{liu2022multi} has harvested an extensive assortment of descriptive molecular texts from PubChem, forging a solid bridge of consistency between these narratives and their corresponding molecule graph. However, to the best of our knowledge, there is currently no multimodal representation learning method that combines molecular structural features with image information.

\section{Methods}
In this paper, we propose a multimodal molecular pre-training framework (MolIG) for predicting molecular properties based on Image and Graph structures, the architecture of which is illustrated in Figure 1.
The framework comprises five components: the Graph Encoder, the Image Encoder, the Graph Non-Linear Projection, the Image Non-Linear Projection, and the Contrastive Learning. 
We will first detail each part of the network architecture, followed by an explanation of inference on downstream tasks.

\subsection{Graph And Image Encoders}

In this section, we will introduce the encoders for the Molecule Graph and the Molecule Image.

\begin{figure*}[!t]%
\centering
\includegraphics[width=1\textwidth]{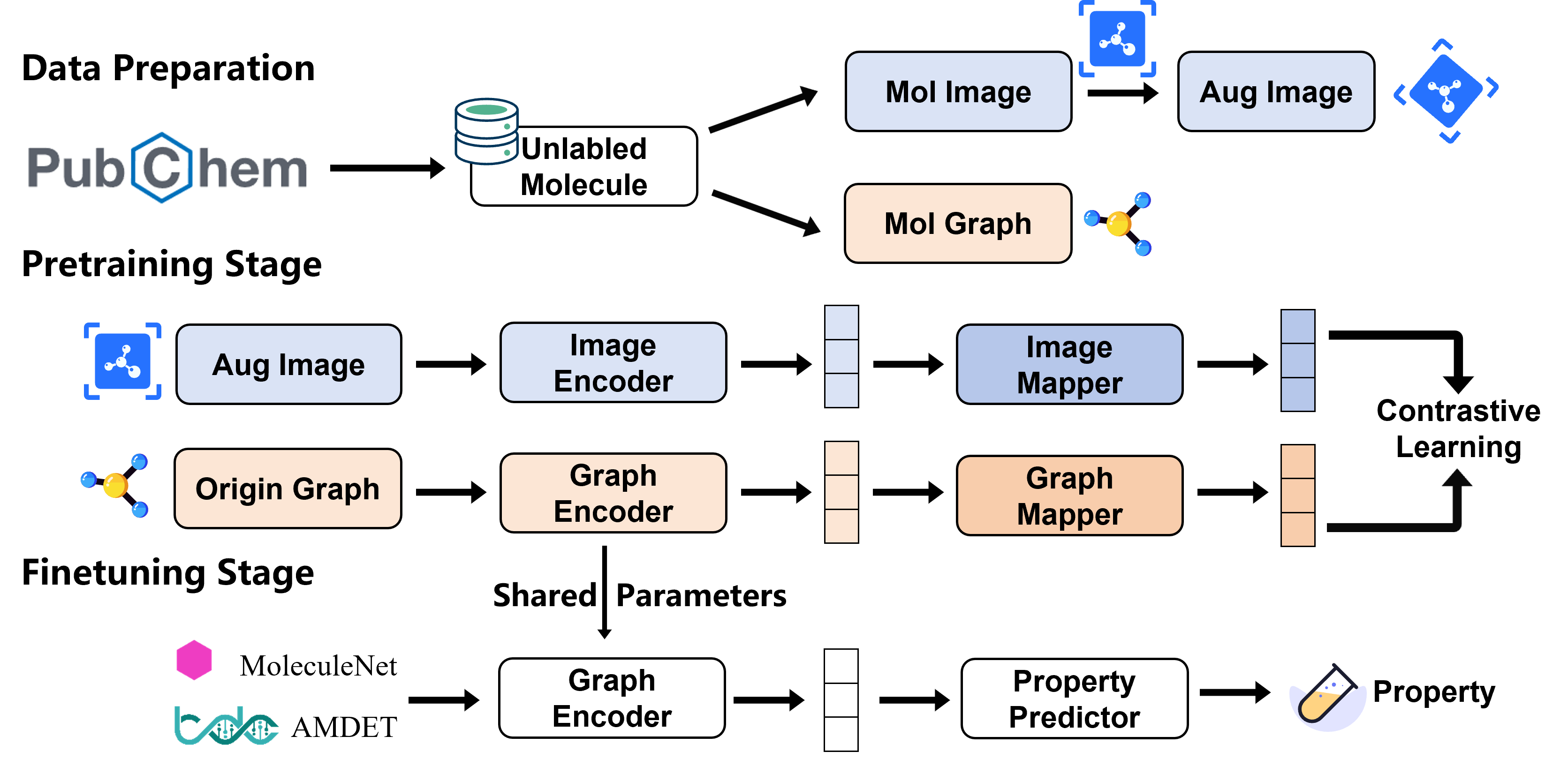}
\caption{Overview of our method: MolIG.}\label{fig1}
\end{figure*}

\subsubsection{Graph Encoder}

A molecule graph $G$ can be defined as $G=(V, E)$, where each node $v\in V$, and each edge $e_{uv} \in E$ signifies the chemical bond between atoms $u$ and $v$.

\begin{equation}
a_v^{\left(k\right)}=AGGREGATE^{\left(k\right)}\left(\left\{h_u^{\left(k-1\right)}:u\in\mathcal{N}\left(v\right)\right\}\right)\label{eq1}
\end{equation}

\begin{equation}
h_v^{\left(k\right)}=COMBINE^{\left(k\right)}\left(h_v^{\left(k-1\right)},a_v^{\left(k\right)}\right)\label{eq2}
\end{equation}

$\mathcal{N}\left(v\right)$ represents the set of all neighbors of node $v$, $h_{v}^{k}$ 
denotes the representation of atom $v$ at the $k$-th layer. After $k$ iterations, $h_{v}^{k}$ is able to capture the information of its $k$-hop neighborhood. The AGGREGATION function collates the information from neighboring nodes of $v$, and the COMBINE function updates the aggregated features. The initial representation $h_{v}^{0}$  is initialized by the node feature $x_{v}$. Notably, in this work, we only employ two types of atomic attributes, namely atom type and chirality. Similarly, for chemical bonds, we use bond type and direction as initial features.

To further extract graph-level features $h_G$, a readout operation integrates the features of all nodes in graph $G$, as illustrated in Equation (3):

\begin{equation}
h_G=READOUT\left(\left\{h_v^{\left(k\right)}:v\in G\right\}\right)\label{eq3}
\end{equation}

In this work, we employ the Graph Isomorphism Network (GIN) \cite{xu2018how} as the GNN Encoder.   However, as molecule graph differ from other types of graph structures, edge information significantly impacts the downstream tasks, which the original GIN does not take into account. In response to this, we follow Hu et al. \cite{hu2020pretraining} to extend node aggregation to $a_v^{\left(k\right)}=\sum_{u\in\mathcal{N}\left(v\right)}\sigma\left(h_u^{\left(k-1\right)}+e_{uv}\right)$, which considers both node and edge information simultaneously. Here, $\sigma$(·) is a non-linear activation function. The readout operation is an average pooling function, used to obtain the graph-level representation for each molecule. The GIN network has 5 million parameters.

\subsubsection{Image Encoder}

For molecule image, we adopt ResNet \cite{he2016deep} as the encoder to extract features from each molecule image.  
In our task, due to the lightweight nature of ResNet-34, we ultimately chose it as the image backbone. It is worth noting that there is a significant discrepancy between molecule image and ImageNet \cite{krizhevsky2017imagenet}, so we do not use ResNet pre-trained on ImageNet. The Resnet network has 22 million parameters. Instead, we trained the image encoder from scratch to accommodate the unique properties of molecule image, a decision that we validated during our experimentation.

\subsection{Graph and Image Projection Heads}
The projection head serves as a pivotal component in the MolIG model, mapping high-dimensional feature vectors extracted from the Graph Encoder and Image Encoder into a lower-dimensional space. Specifically, the projection head can be regarded as a Multilayer Perceptron (MLP), which aims to learn more discriminative representations in the lower-dimensional space for contrastive learning. We follow the nonlinear MLP mentioned in \cite{chen2020simple}, adding a Relu activation function in the hidden layer. Given the substantial disparity between molecule graph and molecule image, we deploy two identical, albeit non-shared-parameter, projection heads. This approach facilitates superior results in downstream tasks of molecular property prediction. The projection head can be formalized as follows:

\begin{equation}
z_{i}^{graph}=Prejection 
 Head_{graph}\left(h_{i}^{graph}\right)\label{eq4}
\end{equation}

\begin{equation}
z_i^{image}=Prejection 
 Head_{image}\left(h_i^{image}\right)\label{eq5}
\end{equation}
Where $PrejectionHead_{graph}$ and $PrejectionHead_{image}$  denote the Non-Linear Projection Heads of the GNN Encoder and Image Encoder, respectively.

\subsection{Contrastive Learning Framework}
Specifically, given a batch of $N$ molecules $\{M_0,M_1\ldots M_{n-1}\}$, we first convert them into Graphs $\{G_0,G_1,G_2\ldots G_{n-1}\}$ and Images $\{I_0,I_1,I_2\ldots I_{n-1}\}$ via Rdkit \cite{landrum2006rdkit}. The molecule image resolution is 224x244. Graphs and Images derived from the same molecule constitute positive samples, whereas graphs from different molecules are considered negative samples. The GNN-based Graph Encoder encodes $G_i$ into $h_i^{graph}$, and the ResNet-based Image Encoder encodes $I_i$ into $h_i^{image}$. However, due to significant differences between the modalities, directly using $h_i^{graph}$ and $h_i^{image}$ as inputs for contrastive loss often results in suboptimal performance in downstream tasks. To address this issue, we design separate mapping heads for the graph and image modalities, respectively mapping the molecule graph and molecule image to the same projection space to compute the contrastive loss, that is, $z_i^{graph}$ and $z_i^{image}$.

Our goal is to maximize the representation similarity of positive sample molecules under Graph and Image modalities. Therefore, under the Graph-Image view, following SimCLR \cite{chen2020simple}, we use the normalized temperature-scaled cross entropy (NT-Xent) as the loss function:

\begin{equation}
\mathcal{L}_{{i},{j}}=-log\frac{exp{\left(sim\left(z_i^{graph},z_j^{image}\right)/\tau\right)}}{\sum_{k=1}^{2N}{1({k}\neq i)}exp{\left(sim\left(z_i^{graph},z_k^{image}\right)/\tau\right)}}\label{eq6}
\end{equation}

Where $\tau$ denotes the temperature coefficient, and we implement the calculation of similarity as $sim(z_i,z_j)=\frac{z_i^Tz_j}{\left.||z_i\right.||_2 · \left.||z_j\right.||_2}$. The final loss is calculated within a mini-batch.

\subsection{Fine-tuning and Downstream Inference}
In this study, we adopt a strategy of pre-training plus fine-tuning to complete the task of drug property prediction. We discard the Image Encoder from the pre-training phase, retaining only the Graph Encoder as the drug molecule encoder, using the pre-trained model to learn the molecular representation of the drug. Subsequently, we fine-tune the pre-trained model on the task of drug property prediction, enabling it to better adapt to the data distribution of the specific task. Specifically, we add only an additional fully connected layer, PredictionHead($w$), after the Graph Encoder, where $w$ represents the parameters of the fully connected layer. In the fine-tuning process, we first input the molecule graph of the drug, $Drug_i$, into the GNN to obtain the latent representation of the molecule, $h_i$, and then forward it to the PredictionHead to make the final property prediction, formalized as follows:

\begin{equation}
\hat{y}=PredictionHead\left(GNN\left(Drug_i\right)\right)\label{eq7}
\end{equation}

\section{Experiments}\label{sec4}
\subsection{Pretraining Dataset and Data Augmentations}
For the pre-training dataset, we randomly select 10 million unlabeled drug-like molecules from the 114 million molecules in the Pubchem database \cite{kim2019pubchem}. We then divide the pre-training dataset into training and validation sets at a ratio of 0.95/0.05.

Data augmentation serves as an effective approach to enhance the generalization capability and robustness of a model. By introducing more data variations during the training process, models can better adapt to the complex scenarios encountered in the real world. Data augmentation has been widely applied in computer vision and multimodal domains. However, for molecule graph, common augmentation techniques such as random atom masking, edge perturbation, or subgraph sampling can disrupt molecular structural information, resulting in learned representations that do not fit well with the data in downstream tasks. Therefore, we do not perform any augmentation operations on molecule graph during the pre-training phase. In contrast, molecule image is sparser compared to real-world images, with over 90\% of their regions filled with zeros, meaning that only a minimal fraction of the pixels are truly relevant to downstream tasks \cite{goh2017chemception}. In light of this, we chose three augmentation strategies during the pre-training phase: (1) RandomHorizontalFlip, (2) Ran-domGrayscale, and (3) RandomRotation. Each strategy has a 25\% probability of being executed. These strategies do not alter the structure of the molecule image and enable the model to learn the invariance introduced by data augmentation.

\begin{table*}[!ht]
    \centering
    \caption{Test performance of different models on six classification benchmarks. The first two models are supervised learning methods and the last eight are self-supervised/pre-training methods. Mean and standard deviation of test ROC-AUC (\%) on each benchmark are reported.}
    \scalebox{1.4}{
    \begin{tabular}{rlllllllc}
    \toprule
        \textbf{Dateset} & \textbf{BBBP} & \textbf{BACE} & \textbf{SIDER} & \textbf{Tox21} & \textbf{HIV} & \textbf{ToxCast} & \textbf{ClinTox} & \textbf{AVG} \\ 
        Molecule & 2,039 & 1,513 & 1,427 & 7,831 & 41,127 & 8,575 & 1,478 & -  \\ 
        Tasks & 1 & 1 & 27 & 12 & 1 & 617 & 2 & -  \\ 
        Split & Scaffold & Scaffold & Scaffold & Scaffold & Scaffold & Scaffold & Scaffold & -  \\ 
    \midrule
        GCN & 71.8(0.1) & 71.6(2.0) & 53.6(3.2) & 70.9(2.6) & 74.0(3.0) & 60.1(1.3) & 62.5(2.8) & 66.3  \\ 
        GIN & 65.8(4.5) & 70.1(5.4) & 57.3(1.6) & 74.0(0.8) & 75.3(1.9) & 62.2(1.9) & 58.0(4.4) & 66.1   \\ 
        Attentive FP  	&  64.3(1.8)  & 	 78.4(0.1)  & 	 60.6(3.2)  	&  76.1(0.5)  	&  75.7(1.4)  	&  63.7(0.2)  & 	 84.7(0.3)  	& 71.9\\
        MFBERT&	71.6(3.3)&	71.6(4.4)&	61.1(7.4)	&63.9(4.8)&	71.1(2.4)&	63.7(8.3)	&77.9(10.2)&	68.7\\
        BARTSmiles&70.9(3.7)&83.2(3.3)&	57.6(6.9)&65.1(5.0)&70.9(2.6)&64.9(8.6)&79.3(7.5)&70.3\\  
        GraphCL	&67.5(3.3)&	68.7(7.8)&	60.1(1.3)&	74.4(0.7)	&75.0(0.4)	& 63.0(0.4)	&   78.9(4.2)&	69.6\\ 
         GraphMVP & 68.5(0.2) & 76.8(1.1) & 62.3(1.6) & 74.5(0.4) & 74.8(1.4) & 62.7(0.1) & 79.0(2.5) & 71.2   \\ 
        3DInfoMax&	69.1(1.0)	&79.4(1.9)&	53.3(3.3)&	74.5(0.7)&	76.1(1.3)&	 63.5(0.8)	 &  59.4(3.2)	&67.9\\
        GROVER & 68.0(1.5) & 79.5(1.1) & 60.7(0.5) & 76.3(0.6) & 77.8(1.4) & 63.4(0.6) & 76.9(1.9) & 71.8   \\ 
        MGSSL	&69.7(0.9)	&79.1(0.9)&	61.8(0.8)	&76.5(0.3)&	78.8(1.2)&	 63.3(0.5)	& 80.7(2.1)&	72.8\\
        MolCLR	&71.6(0.7)&	81.9(1.5)&	59.9(0.9)&	75.0(0.4)&	78.3(0.4)	& 64.7(0.1)	& 81.9(1.2)&	73.3\\
        Mole-Bert & 71.9(1.6) & 80.8(1.4) & 62.8(1.1) & 76.8(0.5) & 78.2(0.8) & 64.3(0.2) & 78.9(3.0) & 73.3   \\ 
        SimSGT&	72.3(0.7)&	83.6(0.8)&	60.6(0.5)&	75.7(0.5)&	77.7(0.8)&	64.1(0.4)&	82.0(2.6)&	73.7\\
        FG-BERT&	70.2(0.9)&	84.5(1.5)&	64.0(0.7)&	\textbf{78.4(0.8)}&	77.4(1.0)&	\textbf{66.3(0.8)}&	83.2(1.6)	&74.8\\
    \midrule
        \textbf{MolIG} & \textbf{73.4(0.7)} & \textbf{84.5(0.2)} & \textbf{66.1(0.5)} & 75.9(0.1) & \textbf{79.8(0.6)} & 64.8(0.1) & \textbf{89.1(2.0)} & \textbf{76.2}   \\ 
    \bottomrule
    \end{tabular}
    }
    \label{table:3moleculenet_result}
\end{table*}

\subsection{Molecule Property Prediction on MoleculeNet}
\subsubsection{Dataset}
To evaluate the performance of MolIG, we fine-tune the pre-trained model on six classification datasets from MoleculeNet \cite{wu2018moleculenet}, including BBBP, BACE, SIDER, Tox21, HIV, ToxCast and ClinTox. 
Compared to random split, scaffold split is a more challenging and realistic method of division, greatly testing the  robustness and generalization capabilities of the model. We use a scaffold split to divide the above six datasets into training/validation/test sets at a ratio of 0.8/0.1/0.1.

\subsubsection{Baseline}
We compare MolIG with state-of-the-art baseline models for molecular property prediction, which includes supervised learning models like GCN \cite{kipf2017semi}, GIN \cite{xu2018how} and Attentive FP\cite{xiong2019pushing}. The rest of the models are pre-trained models.
\begin{itemize}
  \item MFBERT\cite{abdel2022large}: A Transformer-based approach for chemical fingerprint processing, which involves distributed computing, including pre-training and fine-tuning processes.
  \item BARTSmiles\cite{chilingaryan2022bartsmiles}: A self-supervised strategy based on a generative masked language model, training the BART model to learn effective molecular representations.
  \item Mole-Bert\cite{xia2022mole}: A molecular representation learning method that converts atom encodings into discrete codes with chemical significance and combines masked atom modeling with contrastive learning.
  \item SimSGT\cite{liu2024rethinking}: A pre-training method based on Masked Graph Modeling (MGM), utilizing a simple GNN as a tokenizer to learn molecular representations by reconstructing embeddings of subgraphs.
  \item FG-BERT\cite{li2023fg}: A self-supervised method based on BERT, learning molecular representations with domain knowledge by masking functional groups with chemical semantics.
  \item GraphCL \cite{you2020graph} learns molecular representations by maximizing the mutual information of graphs under two different augmentation strategies.
  \item 3DInfoMax\cite{stark20223d} and GraphMVP\cite{liu2022pretraining} introduce 3D molecular information and maximize the representation of molecules in both 2D and 3D.
  \item GROVER \cite{rong2020self} combines GNN and Transformer, learning molecular representations through two pre-training tasks, namely context prediction and Motif prediction.
  \item MGSSL \cite{zhang2021motif} proposes a universal Motif-based generative pre-training framework, which requires GNN to perform topology and label prediction.
  \item GraphMAE \cite{hou2022graphmae} applies MAE \cite{he2022masked} to Graph, learning effective information by reconstructing node features.
  \item MolCLR \cite{wang2022molecular} performs contrastive learning with molecule graph under two identical augmentation strategies.
\end{itemize}

\subsubsection{Result}
In Table 1, we report the average ROC-AUC values and standard deviations of MolIG on seven benchmark datasets. We independently ran MolIG 3 times on each dataset. Notably, MolIG achieve the best performance on five of the datasets and reached the highest average value across all datasets. We attribute the exceptional performance exhibited by MolIG to the effectiveness of the pre-training model in integrating molecule image and graph structure information. These two modalities together provide a rich context for molecular property prediction, thereby improving the accuracy of the predictions.
Specifically, compared to the SOTA model FG-BERT, which only uses a single molecule graph modality, FG-BERT has a significant performance advantage, surpassing FG-BERT in five of the datasets. Particularly on the BBBP dataset, MolIG achieve a performance improvement of 3.2\%.  Especially on the ClinTox dataset, MolIG achieve a performance improvement of 5.9\%.  This demonstrates the superiority of the multimodal strategy adopted by MolIG in excavating the latent knowledge in molecular data.

\subsection{Molecule Property Prediction on ADMET}\label{subsec3}
\subsubsection{Dataset}
ADMET \cite{huang2021therapeutics} is a crucial concept in drug discovery, representing properties such as drug absorption, distribution, metabolism, excretion, and toxicity. Similar to the MoleculeNet Benchmark Group, we fine-tune MolIG on different tasks in the ADMET Benchmark Group. But with the difference that, we use the molecule scaffold division provided   by ADMET to divide each dataset into training, validation, and testing sets in a 0.8/0.1/0.1 ratio. 
\subsubsection{Baseline}\label{subsubsec2}
In this study, we achieve MolIG with three benchmark models: GCN \cite{kipf2017semi}, Attentive FP \cite{xiong2019pushing}, and MolCLR \cite{wang2022molecular}.
\begin{itemize}
  \item GCN \cite{kipf2017semi} and MolCLR \cite{wang2022molecular} are consistent with the models we used in the MoleculeNet Benchmark Group.
  \item Attentive FP\cite{xiong2019pushing} combines the concepts of attention mechanism and molecular fingerprints to capture complex interactions between atoms and bonds in a molecule. It iteratively updates the features of atoms and bonds to generate vector representations of the molecules.
\end{itemize}

%
%
\begin{table*}[t]
\centering
\caption{Test performance of different models on ADMET benchmark group.  Mean result on each benchmark are reported.}
\scalebox{1.4}{
\begin{tabular}{ccccccc}
\toprule
Category & Dataset & Metric & GCN & AttentiveFP & MolCLR & MolIG \\
\midrule
\multirow{6}{*}{Absorption} & Caco2 & MAE & 0.599 & 0.401 & 0.434 & \textbf{0.341} \\
 & HIA & ROC-AUC & 0.936 & \textbf{0.974} & 0.956 & 0.966 \\
 & Pgp & ROC-AUC & 0.895 & 0.892 & 0.861 & \textbf{0.921} \\
 & Bioavailability & ROC-AUC & 0.566 & 0.632 & \textbf{0.738} & 0.725 \\
 & Lipophilicity & MAE & 0.541 & 0.572 & 0.496 & \textbf{0.461} \\
 & Solubility & MAE & 0.907 & 0.776 & 0.776 & \textbf{0.753} \\
\midrule
\multirow{2}{*}{Distribution} & PPBR & MAE & 10.194 & 9.373 & 9.196 & \textbf{8.499} \\
 & VDss & Spearman & 0.457 & 0.241 & 0.547 & \textbf{0.717} \\
\midrule
\multirow{6}{*}{Metabolism} & CYP2D6 inhibition & PR-AUC & 0.616 & 0.646 & 0.686 & \textbf{0.693} \\
 & CYP3A4 inhibition & PR-AUC & 0.840 & \textbf{0.851} & 0.840 & 0.841 \\
 & CYP2C9 inhibition & PR-AUC & 0.735 & 0.749 & 0.781 & \textbf{0.802} \\
 & CYP2D6 substrate & PR-AUC & 0.617 & 0.547 & 0.486 & \textbf{0.702} \\
 & CYP3A4 substrate & ROC-AUC & 0.590 & 0.576 & 0.543 & \textbf{0.655} \\
 & CYP2C9 substrate & PR-AUC & 0.344 & 0.375 & 0.353 & \textbf{0.400} \\
\midrule
\multirow{3}{*}{Excretion} & Half life & Spearman & 0.239 & 0.085 & 0.133 & \textbf{0.357} \\
 & Clearance microsome & Spearman & 0.532 & 0.365 & 0.525 & \textbf{0.561} \\
 & Clearance hepatocyte & Spearman & 0.366 & 0.289 & 0.386 & \textbf{0.387} \\
\midrule
\multirow{4}{*}{Toxicity} & hERG & ROC-AUC & 0.738 & 0.825 & 0.821 & \textbf{0.831} \\
 & Ames & ROC-AUC & 0.818 & 0.814 & \textbf{0.887} & 0.883 \\
 & DILI & ROC-AUC & 0.856 & 0.886 & 0.871 & \textbf{0.921} \\
 & LD50 & MAE & 0.649 & 0.678 & 0.435 & \textbf{0.428} \\
\bottomrule
\end{tabular}
}
\end{table*}

\subsubsection{Result}\label{subsubsec3}
In Table 2, we report the average ROC-AUC values for MolIG on 21 benchmark datasets in ADMET. Compared to the MoleculeNet Benchmark Group, the ADMET Benchmark Group focuses more on the behavioral characteristics of drug molecules within biological entities, and therefore, contains relatively more complex molecular structures. The experimental results show that MolIG achieves the best performance on 17 out of the 21 datasets. Except for the Bioavailability and Ames datasets, our model outperformed MolCLR, which only uses Graph modality. Particularly, in the Pgp dataset, we improve the performance by nearly 3\% compared to the best baseline. In the VDss dataset, the performance was improved by nearly 17\%. In the CYP2D6 substrate, MolIG achieves a performance improvement of nearly 9\%. These results suggest that by integrating molecule image information into the model, MolIG has a more significant advantage in predicting more complex properties within biological entities. 
The outstanding performance of MolIG on the ADMET benchmark datasets confirms the effectiveness of the multimodal pre-training framework in the field of molecular property prediction. This research provides useful insights for further exploring multimodal learning strategies to improve the accuracy of drug molecule property prediction in the future.

\subsection{Ablation Study}\label{subsec4}
This ablation study aims to evaluate whether pre-training strategies and data augmentation strategies can help the model achieve better performance in downstream tasks. As shown in Figure 2, MolIG, which utilizes both pre-training and data augmentation strategies (represented by the grey bar), performs best among all model architectures. Models without pre-training (represented by 'w/o pretrain') perform worst in all tasks. Pre-training strategies without data augmentation (represented by 'w/o img aug') perform second best, yet they show a significant performance increase compared to models without any pre-training. Excluding either of these two components can easily lead to a decrease in performance.

Compared to models completely without pre-training, MolIG improves the ROC-AUC metric by 13.0\% on the ClinTox dataset, by 7.8\% on the SIDER dataset, and an average of 5.9\% across all six datasets. The 'w/o img aug' strategy, although not using data augmentation, still shows a noticeable improvement, indicating the effectiveness of our pre-training strategy. MolIG improves across all six datasets, suggesting that the data augmentation strategy further enhances  robustness and generalization capabilities of the model.

In conclusion,  multimodal pre-training strategy of MolIG can learn high-level semantic information from a large amount of unlabeled molecular data, enabling the model to learn more discriminative representations. Meanwhile, the data augmentation strategy enhances  robustness and generalization capabilities by perturbing the image modality without changing the semantic information and effectively transfers to relevant biological tasks for molecular property prediction.

\begin{figure}[!t]%
\centering
\includegraphics[width=0.5\textwidth]{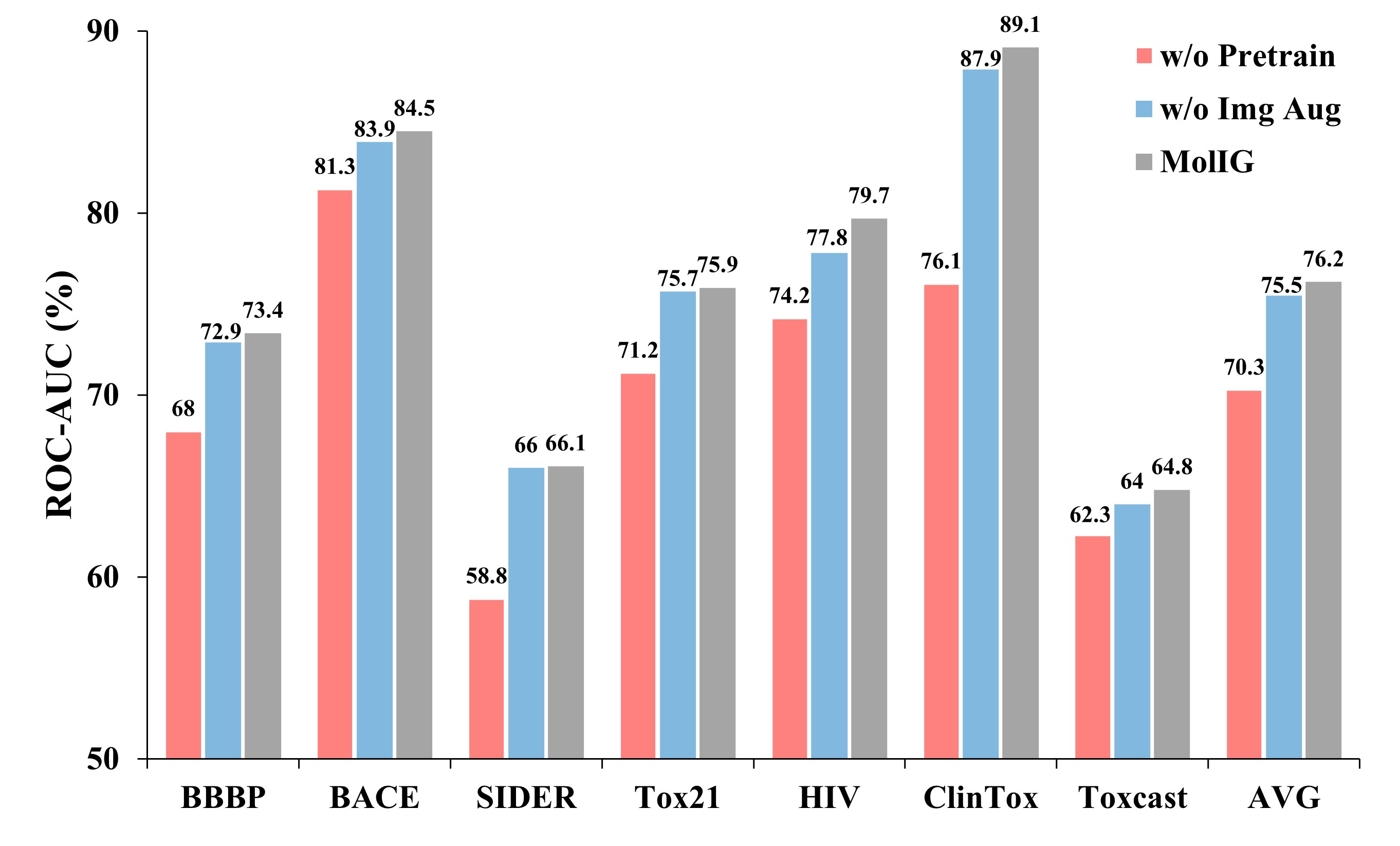}
\caption{Ablation Study: Impact of Pre-training and Data Augmentation Strategies on the Performance of MolIG on Six Classification Datasets in MoleculeNet.}\label{fig2}
\end{figure}

\subsection{Parameter Experiment}\label{subsec5}
A critical hyperparameter in the MolIG model is the temperature scaling used in contrastive learning, which adjusts the  sensitivity of the model to the similarity scores of positive and negative samples. As shown in Table 3, to investigate the influence of different temperature coefficients on the performance of downstream tasks, we adopt three different temperature coefficients during the pre-training phase of the model, namely 0.05, 0.1, and 0.5, and conduct experiments on six classification datasets on MoleculeNet, reporting the average ROC-AUC values under these three temperature coefficients. Moreover, Figure 3 illustrates the specific performance of different temperature coefficients on these datasets.

From Table 3 and Figure 3, it can be observed that when the temperature coefficient is 0.1, the performance of the model on downstream tasks reaches its peak. We posit that a lower temperature coefficient assists the model in distinguishing negative samples more effectively. However, when the temperature coefficient is 0.05, its average performance on downstream tasks is 2\% lower than that of the model with a temperature coefficient of 0.1. This may be due to the sparsity of molecule image causing the model to transmit redundant information from the image modality to the GNN encoder. A higher temperature coefficient, on the other hand, simplifies the model training process, preventing it from learning meaningful representations. We have observed that when the temperature coefficient is set to 0.5, the model struggles to converge, leading to unstable training. Moreover, it reduces the discriminative ability between different molecules, resulting in ambiguous predictions. This could significantly impact its performance in downstream tasks. 

\begin{table}[]
\centering
\caption{The Impact of Temperature ($\tau$) on the Loss of MolIG: A Report on the Average Performance across Six Classification Datasets.}
\begin{tabular}{lllll}
\toprule
Temperature ($\tau$) & 0.05 & 0.1  & 0.5  &  \\
\midrule
ROC-AUC (\%)      & 74.2 & 76.2 & 73.4 &  \\
\hline
\end{tabular}
\end{table}

\begin{figure}[!t]%
\centering
\includegraphics[width=0.5\textwidth]{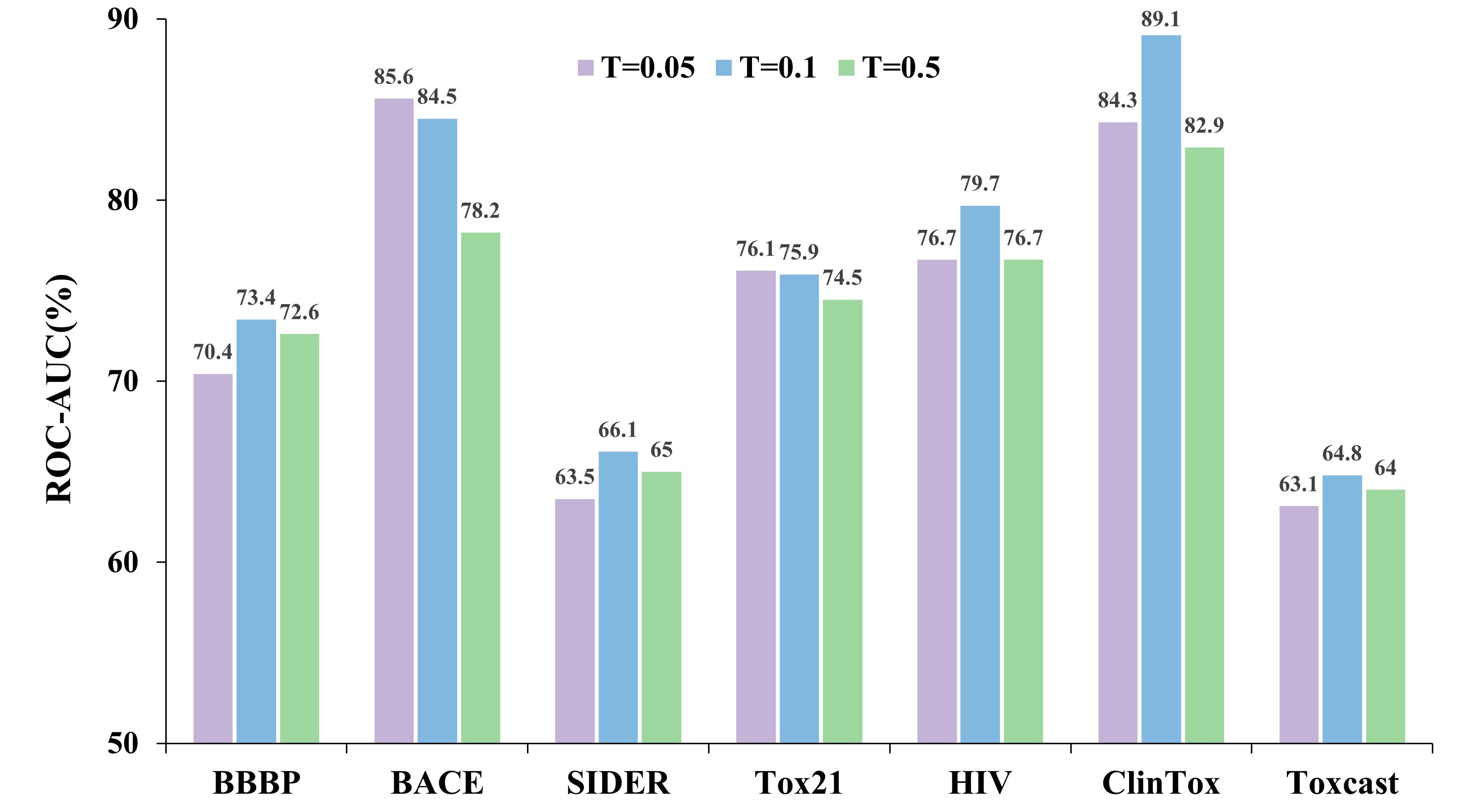}
\caption{Performance of MolIG on six classification datasets in MoleculeNet under different temperature coefficients.}\label{fig3}
\end{figure}

\subsection{Molecule Retrieval via MolIG}\label{subsec2}
To further investigate the representational capabilities of the MolIG method, we evaluated its performance in the field of molecular similarity search. In the experimental design, we initially randomly selected 100,000 molecules from the PubChem database as the query set. For each query molecule, we obtained its molecular representation using MolIG and subsequently calculated its cosine similarity with the representations of the remaining molecules in the database.
\begin{figure*}[t]
\centering
\includegraphics[width=\linewidth]{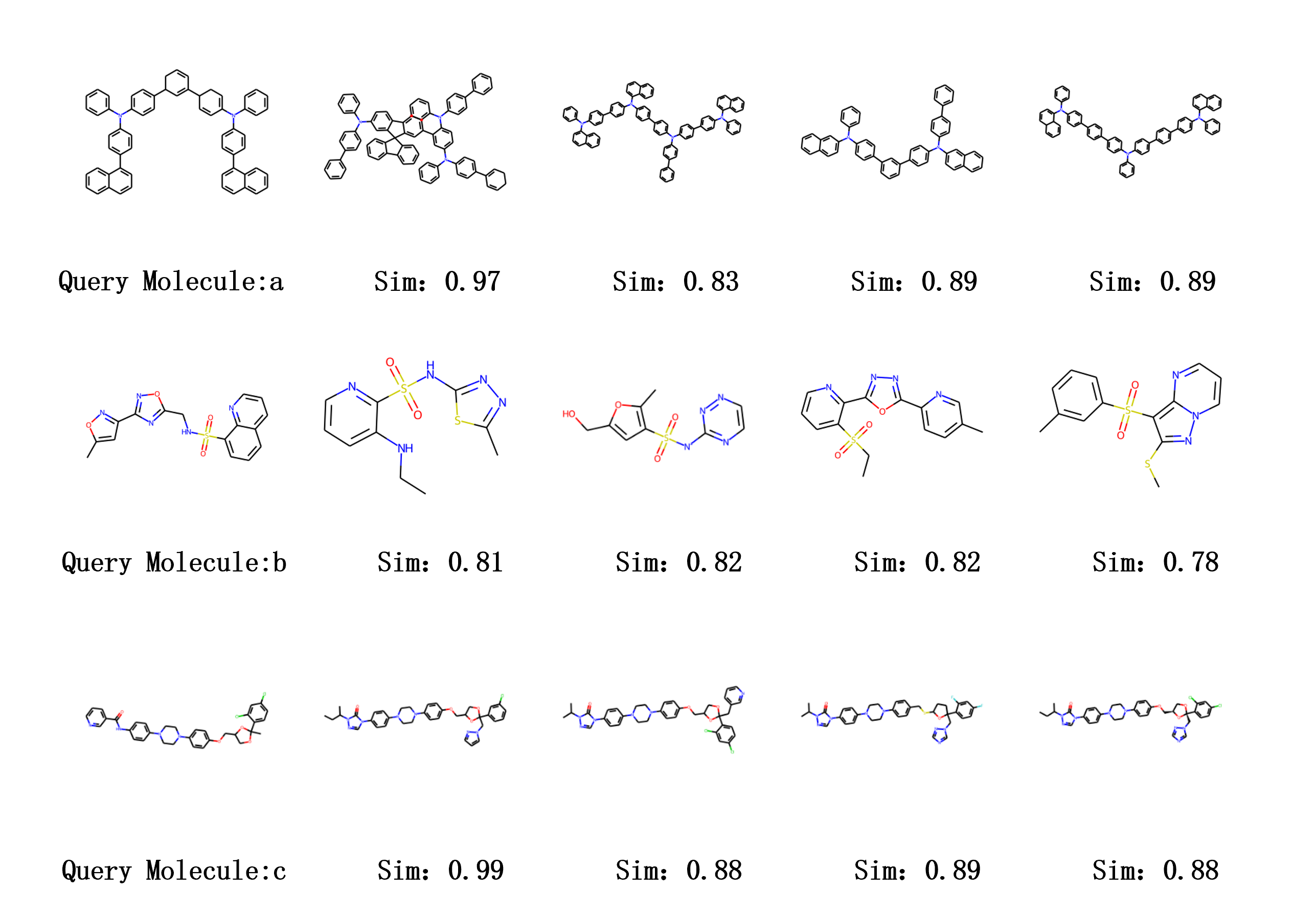}
\caption{Query results}\label{fig4}
\end{figure*}

Based on the computed similarity scores, the experimental results showcase several instances of molecules with the highest similarity to the query molecules, presented in Figures 4. For instance, concerning query molecule a, characterized by a structure composed of multiple benzene rings, MolIG successfully retrieved molecules with similar arrangements of benzene rings and a central nitrogen atom. This result validates the effectiveness of MolIG in capturing atomic-level structural information. For query molecule b, MolIG accurately identified its key functional group O=S=O, demonstrating the model's capability in recognizing molecular functional groups. In the query experiment for molecule c, the molecules retrieved by MolIG commonly contain halogen elements such as fluorine or chlorine and exhibit toxic properties, indicating MolIG's ability to understand molecular properties from a semantic perspective. In summary, these experimental results indicate that MolIG can not only precisely capture molecular structural details but also comprehend higher-order molecular semantic information, such as the features of functional groups and the potential attributes of molecules.

\section{Implemention Ddtails}\label{sec5}
Our pre-training model consists of a Graph Isomorphism Network (GIN) with 5 layers and 300 hidden dimensions and a residual convolutional neural network (ResNet-34) with 22M parameters. We pre-train the model for 50 epochs using a batch size of 512 on 1 RTX 3090 GPU. We use the Adam optimizer, the ResNet learning rate is 0.01 and the GIN learning rate is 0.0005. The temperature coefficient for Contrastive loss is 0.01. The weight decay is 1e-5. We take image with resolution of 224×224. 

\section{Conclusions}\label{sec5}
Molecular property prediction plays a crucial role in drug discovery. While previous models for predicting molecular properties have achieved considerable success, the use of a single modality of information often limits their predictive performance. In this study, we propose a novel multi-modal molecular pre-training framework MolIG which learns molecular representations by maximizing the consistency between the features of the Graph and Image modalities during the pre-training phase. The empirical evaluations on drug discovery tasks such as MoleculeNet and ADMET demonstrate that MolIG outperforms the current state-of-the-art baseline models. Notably, MolIG not only extracts structural features of molecules but also captures higher-order semantic information, which can be transferred to biologically relevant tasks in molecular property prediction. Furthermore, our model currently only considers information from two modalities. Incorporating SMILES and 3D information into the framework will be the subject of our future research.

\section*{Acknowledgments}
This work was supported by National Key R\&D Program of China (2019YFC1711000) and Collaborative Innovation Center of 
Novel Software Technology and Industrialization.

\bibliographystyle{unsrt}
\bibliography{reference}
\end{document}